# Assessing Shape Bias Property of Convolutional Neural Networks

Hossein Hosseini    Baicen Xiao    Mayoore Jaiswal    Radha Poovendran
Network Security Lab (NSL)
Department of Electrical Engineering, University of Washington, Seattle, WA

## Abstract

*It is known that humans display "shape bias" when classifying new items, i.e., they prefer to categorize objects based on their shape rather than color. Convolutional Neural Networks (CNNs) are also designed to take into account the spatial structure of image data. In fact, experiments on image datasets, consisting of triples of a probe image, a shape-match and a color-match, have shown that one-shot learning models display shape bias as well.*

*In this paper, we examine the shape bias property of CNNs. In order to conduct large scale experiments, we propose using the model accuracy on images with reversed brightness as a metric to evaluate the shape bias property. Such images, called negative images, contain objects that have the same shape as original images, but with different colors. Through extensive systematic experiments, we investigate the role of different factors, such as training data, model architecture, initialization and regularization techniques, on the shape bias property of CNNs. We show that it is possible to design different CNNs that achieve similar accuracy on original images, but perform significantly different on negative images, suggesting that CNNs do not intrinsically display shape bias. We then show that CNNs are able to learn and generalize the structures, when the model is properly initialized or data is properly augmented, and if batch normalization is used.*

## 1. Introduction

Recently, it has been shown that state-of-the-art Convolutional Neural Networks (CNNs) trained with stochastic gradient descent (SGD) have enough capacity to "memorize" the training data, even when training images or labels are randomized [1]. Memorization is often associated with overfitting training data and, hence, large generalization error, defined as the difference between training error and test error. Yet, CNNs are known to be able to generalize well to images that are similarly distributed as training data.

For an image classifier, the space of possible inputs is much larger than the size of training data. Hence, models with identical performance on samples that are similarly distributed as training images can perform qualitatively differently on other distributions. Of course, most of possible inputs are random images that the model is not expected to recognize. Therefore, it remains to be determined what distributions we expect the model to generalize to and how different training methods affect generalization capability.

The distributions that we are interested in can be determined according to the human cognitive system. It is known that humans display "shape bias" when assigning a name to new items [2], i.e., they weight shape more heavily than other dimensions of perceptual similarity, such as size or texture. In [3], the authors studied shape bias property in CNNs by performing experiments on small datasets, in which images were arranged in triples of a probe, a shape-match and a color-match. It was shown that state-of-the-art one-shot learning models display shape bias as well, although the magnitude of bias varies greatly with different initializations and also fluctuates throughout training.

In order to conduct larger scale experiments, we propose to train and test CNNs with specifically designed images that represent the same object, but with different colors. One simple way of generating such images is applying image complementing transformation. The transformed image, called *negative image*, is an image with reversed brightness. Image complementing maintains the structure (e.g., edges) and semantics of images, as negative images are often easily recognizable by humans. Figure 1 shows representative samples of original and negative images of MNIST [4], notMNIST [5] and CIFAR10 [6] datasets.

We then examine the shape bias property of CNNs, by training and testing them on negative images. Through extensive systematic experiments on MNIST, notMNIST and CIFAR10 datasets, we assess the role of different factors, such as training data, network architecture, initialization method and regularization techniques, on the ability of model in generalizing the shapes. Our contributions are summarized in the following.

- We show that it is possible to design different CNNs that achieve similar accuracy on original images, but perform significantly differently on negative images.

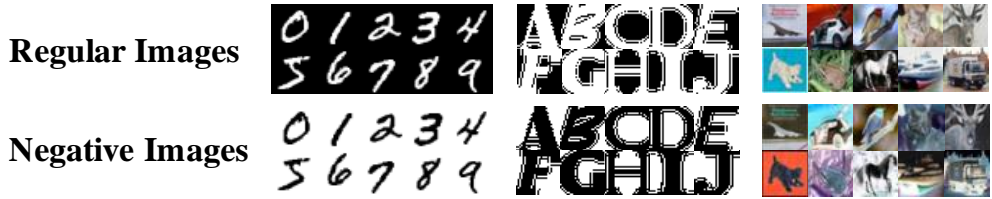

Figure 1: Examples of regular images and their corresponding negative images of datasets MNIST, notMNIST and CIFAR10.

For instance, we designed three CNNs that achieve 0%, 28.3% and 99.5% accuracy on MNIST negative images, while yielding the same accuracy of 99.5% on original images. The results suggest that CNNs do not intrinsically require shape bias property to achieve high accuracy on test data.

- We then show that, when negative images of some classes are included in training data, CNNs can correctly recognize negative images of other classes. This is however true, only if the model is trained with batch normalization. For instance, we trained the CNN with all MNIST regular images and also negative images of 9 classes, and tested it on negative images of the excluded class. The test accuracy is 0% and 97.7% respectively without and with batch normalization. Hence, CNNs can indeed learn to be invariant to color, when properly tuned.

- We also investigate the effect of augmenting training data with original and negative images of an unrelated dataset. We show that such data augmentation also significantly improves generalizability to negative images. For instance, the CNN accuracy on negative images of MNIST training data increases from 29.1% to 99.8%, when MNIST training data is augmented with original and negative images of notMNIST dataset. The results suggest that augmenting training data with unrelated datasets that even have unrelated labels can help the model to learn better image representations.

- We examine the role of initialization and demonstrate that a different initialization can significantly affect the generalization. Specifically, a model that is initialized by training with original and negative images of a dataset shows shape bias on another dataset as well. Moreover, the shape bias property does not fade away after fine-tuning only with original images of the second dataset. For instance, when initialized by a CNN that is trained on original and negative images of notMNIST dataset, the accuracy on MNIST negative images increases from 29.1% to 95.4%.

- We finally show that CNNs do not learn the shape bias property from a dataset that lacks any structure, e.g., a dataset with random images. Moreover, a model with shape bias cannot generalize this property to random images, i.e., it does not perform similarly on random images and their negatives. The results imply that, although CNNs can fit any training data, they only learn and generalize the structures.

The rest of this paper is organized as follows. Section 2 reviews related literature and Section 3 presents setup for experiments. In Section 4, we investigate whether CNNs display shape bias by design. In Section 5, we examine whether CNNs can learn to be invariant to color. Section 6 assesses the effect of initialization and data augmentation on shape bias of CNNs and Section 7 concludes the paper.

## 2. Related Works

Recently, there has been an interest in exploring various aspects of the generalization of CNNs by both theoretical and empirical analysis [1, 7–11]. It has been shown that CNNs trained with SGD are capable of memorizing the training data, contradicting their low generalization error [1]. In [7], however, the authors argued that networks trained with SGD learn structures before memorizing. This follows the works suggesting that SGD generally results in simpler models [12, 13].

Deep neural networks are known to be capable of approximating any measurable function given sufficient capacity [14, 15]. These works determine the set of hypotheses a model can express, but do not specify what hypothesis can be reached by training a network on a dataset using a particular method [16, 17]. In this paper, we assess the capability of CNNs in generalizing shapes when color information is lost, a property known as shape bias [2, 3].

From the practical perspective, it is important to determine how CNNs perform on related distributions as of the training data. This problem has been studied in literature of transfer learning [18, 19] and domain adaptation [20–23]. The goal is to use models and features learned on one dataset/domain for another dataset/domain with minimal fine-tuning [18]. In this paper, we examine transferability of features from original to negative images. To the best of our knowledge, generalizing to negative images is not studied in transfer learning or domain adaptation literature.

Figure 2: Illustration of the three experiments on MNIST. The figure shows one image and the corresponding label as a representative of each class. For example, in experiment 3, regular training images of digit 0 and negative training images of digit 9 are mapped to class zero, and the accuracy on negative test images with correct labels is 0% and accuracy on negative test images with modified labels is 99.5%. In all cases, accuracy on training data is 100%.

## 3. Setup for Simulations

Experiments are performed on image datasets MNIST [4], CIFAR10 [6] and notMNIST [5]. MNIST is a dataset of handwritten digits. CIFAR10 dataset consists of natural color images in 10 classes. notMNIST is a dataset similar to MNIST with 10 classes of letters $A$-$J$ taken from different fonts. notMNIST dataset contains more than $500{,}000$ images, from which we randomly select $60{,}000$ images for training and validation.

In [3], the authors studied the shape bias property in CNNs by performing experiments on images that were arranged in triples of a probe, a shape-match and a color-match. Images were chosen from a cognitive psychology probe data and also a real-world dataset consisting of 150 and 90 images, respectively. In order to conduct larger scale experiments, we propose to train and test CNNs with negative images. A negative image is defined as the complement of the original image, in which light pixels appear dark and vice versa. Let $X$ be an image and $X_{i,n} \in [0,1]$ be the $i$-th pixel in the $n$-th color channel. The negative image is defined as $X^*$, where $X^*_{i,n} = 1 - X_{i,n}$.

Image complementing is a simple transformation that preserves the shape and semantics of images, as negative images are often easily recognizable by humans. In fact, it has been shown that human accuracy on negative images of German Traffic Sign Recognition Benchmark decreases only about 1% compared to original images [24]. Moreover, since in MNIST, notMNIST and CIFAR10 datasets, classes are very distinct, image complementing is unlikely to change the ground-truth label. In the rest of the paper, we refer to original images as regular images.

We consider a small version of VGG-16 model [25], namely sVGG, with 6 convolutional layers followed by two fully connected layers. Similar to VGG-16, we use $3 \times 3$ convolution kernels and $2 \times 2$ max-pooling.[1] For MNIST, we also consider multi-layer perceptrons (MLPs) with 1 or 2 layers, ReLU activation function and 1000 hidden nodes per layer. Models are trained using SGD and, unless otherwise stated, with batch normalization in all layers. In all experiments, 20% of training data are held out and used for validation. We stop training, when validation accuracy is the highest and training accuracy is 100%. In cases that training accuracy does not reach 100%, models are trained for 500 epochs. When validation accuracy is not meaningful (e.g., training with random images), we report test accuracy of the last epoch. Results are averaged over five experiments.

## 4. Testing on Negative Images

In this section, we consider the following questions: 1) Do CNNs display shape bias by designs? and 2) Do they need to achieve shape bias at all in order to yield high accuracy on test data? To answer these questions, we examine the performance of CNNs on negative images by designing three experiments on MNIST and CIFAR10 datasets. The experiments on MNIST are illustrated in Figure 2 and the results on CIFAR10 are provided in Table 1.

In the first experiment, we train the sVGG model only with regular images. As can be seen in Figure 2 and Table 1,

---

[1] The sVGG structure is as follows: (conv $3 \times 3 \times 16$, ReLU, conv $3 \times 3 \times 16$, ReLU, max pool $2 \times 2$, conv $3 \times 3 \times 32$, ReLU, conv $3 \times 3 \times 32$, ReLU, max pool $2 \times 2$, conv $3 \times 3 \times 48$, ReLU, conv $3 \times 3 \times 48$, ReLU, max pool $2 \times 2$, FC-128, ReLU, FC-128, ReLU, FC-10, softmax).

the accuracy on negative test images is significantly lower then the accuracy on regular images. We repeated the experiment without using batch normalization in training, and obtained 12.2% and 21.4% accuracy respectively on MNIST and CIFAR10 negative test images, which signifies the importance of batch normalization in generalization. We will explore the role of batch normalization more in next section. We also performed the experiment with MNIST and using a softmax classifier and 1- and 2-layer MLP models and achieved 0% and around 8% accuracy on negative images, respectively for softmax classifier and MLPs. Hence, regarding generalization to images with similar shapes and different colors, CNNs (with batch normalization) indeed perform better than MLPs.

In the second experiment, we train the models with both regular and negative images. In this case, the accuracy of CNN model on negative images is similar to the accuracy on regular images. We again did the experiment with MNIST and using a softmax classifier and the MLP models. For the softmax classifier, the accuracy even on training data is about 15%, since the data is not linearly separable. The two MLP models, however, yield high accuracy on both regular and negative images.

In the third experiment, we train the models with both regular and negative images, but the labels of negative images are changed to $(i+1) \bmod 10$, where $i$ is the ground-truth label. In this case, the model cannot classify images solely based on the shape pattern, since training images in separate classes have the same shapes. Therefore, while the model can certainly achieve 100% training accuracy, the question is whether it can generalize to regular images and also to negative images with modified labels. The answer is yes for both MNIST and CIFAR10 datasets. Specifically, for MNIST, the accuracy on regular and negative test images (with original labels) is 99.5% and 0%, respectively, and the accuracy on negative images with modified labels is 99.5%. We did the experiment with MNIST and MLP classifiers and obtained similar results.

The results on CIFAR10 is similar to MNIST. Note that, unlike MNIST, CIFAR10 consists of natural images. Therefore, it is interesting that the CNN model generalizes to both regular test images and negative test images with modified labels. For instance, negative images of "automobile" class can be classified as "bird," yet the model would achieve high accuracy on both classes of "automobile" and "bird." Specifically, compared with the case where the model was only trained on regular images, the accuracy on regular test images decreases only about 4%, with most of newly misclassified images are labeled as $(i + 1) \bmod 10$. Similar result holds for negative test images, i.e., only about 4% of negative test images are classified as their true label. [2]

---

[2]The possibility for the model to yield high accuracy both on regular test images and on negative test images with modified labels can be at-

Table 1: Accuracy of sVGG model on regular and negative images of CIFAR10 test data. In experiment 1, the model is only trained with regular images. In experiment 2, the model is trained with both regular and negative images. In experiment 3, the model is also trained with both regular and negative images, but the labels of negative images are changed to $(i+1) \bmod 10$, where $i$ is the ground-truth label. In all cases, accuracy on training data is 100%.

| Test Data | Experiment 1 | Experiment 2 | Experiment 3 |
|---|---|---|---|
| Regular images | 79.7% | 78.3% | 75.4% |
| Negative images | 38.7% | 78.5% | 4.2% |
| Negative images with modified labels | NA | NA | 75.7% |

To conclude, we designed three CNNs that perform similarly on regular images, but very differently on negative images. The results show that CNNs do not intrinsically display shape bias. Specifically, in the second and third experiments, we guided the models to respectively weight shape and color more, and yet they can achieve similar accuracy on regular test images. The results also suggest that accuracy on images that are distributed similarly as training data is not representative of the behavior of machine learning models in the wild. That is, models with identical performance on regular images can behave qualitatively differently on a distinct yet related distribution.

## 5. Training with Negative Images

In this section, we investigate whether CNNs can learn to be "invariant to color." That is, if negative images of some classes are included in training data, does the CNN model correctly recognize negative images of other classes? To answer this question, we first train the sVGG model with all regular images and also negative images of some classes, and test it on negative images of excluded classes. We then extend the experiments by combining two datasets and examine whether the CNN model can transfer features learned on one dataset to another. The latter experiment is conducted on MNIST and notMNIST datasets, since they contain images of similar types.

### 5.1. Experiments on One Dataset

For both MNIST and CIFAR10 datasets, we train the sVGG model with all regular images and negative images of 9 classes of training data. We then test the model on negative training images of the excluded class. We do the experiment on all 10 classes and report the average accuracy. Figure 3 illustrates the experiment.

tributed to the dataset bias discussed in [26].

|  | Training images | Test images | Accuracy | |
|---|---|---|---|---|
| Labels: | 0 1 2 3 4 5 6 7 8 9 | 9 | with BN | w/o BN |
| MNIST | 0123456789 / 012345678 | 9 | 97.7% | 0% |
| CIFAR10 | (images) | (image) | 75.8% | 15.6% |

Figure 3: An illustration of partially training the model with negative images. BN is batch normalization. We train the CNN with all regular images and also negative images of 9 classes, and test it on the negative images of the excluded class. The experiment is conducted on all classes and the average accuracy is reported. In all cases, accuracy on training data is 100%.

Recall from Figure 2 that, for MNIST dataset, when the model is only trained with regular images, the accuracy on negative images is 28.3%. By including negative images of 9 classes into the training data, the accuracy on negative images of the excluded class jumps to 97.7%. We repeated the experiment with MNIST and with 1- and 2-layer MLPs. For MLP models, when including negative images of 9 classes in training data, the accuracy on negative images of the excluded class is 0%. Therefore, unlike MLPs which classify inputs based on raw pixel intensities, CNNs are able to classify objects based on their shapes.

To gain an insight into intermediate feature layers, we visualized the outputs of convolutional layers of the sVGG model trained on MNIST. Two cases are considered: 1) training only with regular images, and 2) training with all regular images and also negative images of classes 0 to 8. The trained models are then tested with a regular and a negative image of digit 9. Figure 4 shows outputs of second convolutional layer with the most number of active neurons. As can be seen, the outputs of the first model are different for regular and negative images. However, for the second model, the first two convolutional layers already provided the invariance to color.

The results on CIFAR10 is similar to MNIST. The accuracy of sVGG on negative images of the excluded class jumps from 38.7% to 75.8%, when negative images of 9 classes are included in training data. Essentially, by including some of negative images in training data, the model learns to weight edge patterns more then the color information. Hence, it can generalize significantly better to other negative images as well.

**Role of regularization and batch normalization.** We examined the effect of $L_2$ and Dropout [27] regularizations on model ability to generalize to negative images of the excluded class. Our experimental results show that regularizing the model slows down and stabilizes the training process, but does not affect the generalization.

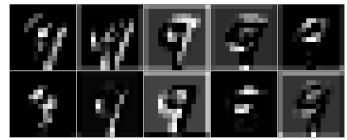

(a) CNN model only trained with regular images.

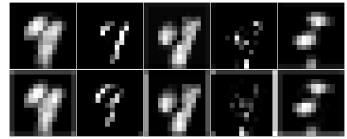

(b) CNN model trained with all regular images and also negative images of classes 0 to 8.

Figure 4: Visualizations of outputs of second convolutional layer. Out of 16 outputs, five of them with most number of active neurons are shown. In each figure, top and bottom rows show outputs on a regular image of digit 9 and its negative, respectively. As can be seen, by simultaneously training with regular and negative images, the model becomes invariant to color.

We also evaluated the role of batch normalization [28] and found that it fundamentally impacts the model behavior on negative images. Essentially, in our experiment, since the model is trained and tested on different distributions, it must deal with the covariate shift problem [29]. Batch normalization is specifically designed to reduce the internal covariate shift of deep neural networks, by fixing the means and variances of layer inputs within each mini-batch. Table 2 summarizes the results of the role of batch normalization on model capability to generalize shapes. As can be seen, CNNs display shape bias only when batch normalization is used. This behavior was consistent across all of our experiments that involved training and testing on different distributions. In the rest of the paper, we only report the experimental results with batch normalization.

Table 2: Accuracy of different models with and without batch normalization (BN). In case 1, models are trained with regular images and tested on negative images. In case 2, models are trained with all regular images and also negative images of 9 classes, and tested on negative images of the excluded class.

| Dataset | Model | Case 1 | | Case 2 | |
|---|---|---|---|---|---|
| | | w/o BN | with BN | w/o BN | with BN |
| MNIST | MLP | 8% | 8% | 0% | 0% |
| | CNN | 12.2% | 28.3% | 0% | 97.7% |
| CIFAR10 | CNN | 21.4% | 38.7% | 15.6% | 75.8% |

**Role of diversity of training data.** We now examine whether enhancing the diversity of negative images improves the generalization. For this, we train the sVGG model with all MNIST regular images and also $10,000$ negative images, in two cases: 1) negative images chosen from classes 0 to 3, and 2) negative images chosen from classes 0 to 7. Figure 5 illustrates the results. As can be seen, increasing the number of classes while keeping number of negative images the same improves the generalization performance. Essentially, more number of classes enhances the diversity of negative images, and hence helps the model to better recognize negative images of unseen classes.

## 5.2. Experiments on Two Datasets

We extend the experiments by combining two datasets and training on negative images of one of them. We train the sVGG model simultaneously with regular and negative images of notMNIST and regular images of MNIST. The model has 20 number of classes, i.e., each class in the two datasets is assigned to a unique label. The experiment is illustrated in Figure 6a. When tested on MNIST negative images, the model accuracy reaches $99.8\%$, almost as if it was also trained on them. In essence, the model learns the class representations using MNIST images and improves its basic understanding of objects, e.g., the shape bias property, using regular and negative images of notMNIST.

**Accuracy versus different number of negative images.** We also investigate whether any number of notMNIST negative images will always lead to better generalization to MNIST negative images. Figure 7 shows the accuracy on MNIST negative images versus different number of notMNIST regular and negative images included in training data. As can be seen, training with few negative images causes the accuracy to decrease. As an example, recall from Figure 2 that, when trained with regular images, the accuracy on MNIST negative images is $28.3\%$. By including only 10 to 100 notMNIST negative images in training data, the model accuracy on MNIST negative images drops to $0\%$.

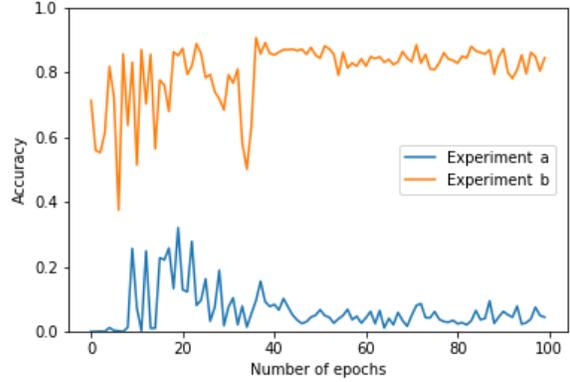

Figure 5: Evaluating the effect of diversity of negative images in training data on accuracy on negative images of excluded classes. The sVGG model is trained with all MNIST regular images and $10,000$ negative images. In experiment a, negative images are chosen from classes 0 to 3, while in experiment b, negative images are chosen from classes 0 to 7. As can be seen, more number of classes with same number of negative images improves generalizability to negative images of excluded classes.

The accuracy, however, reaches $99\%$ when the model is trained with 10000 notMNIST regular and negatives images. This implies that the CNN model learns to classify objects by their shapes, only when it is trained with a large enough number of regular and negative image pairs.

## 6. Role of Data Augmentation and Initialization on Shape Bias

In this section, we examine the transferability of features from one dataset to another, by studying the effect of augmenting MNIST dataset with notMNIST dataset and the role of initialization on model generalizability. We also investigate whether the features can be transferred to or from a dataset with random images.

### 6.1. Role of Data Augmentation

Similar to Section 5.2, we train the sVGG model with regular and negative images of notMNIST and regular images of MNIST, but with 10 number of classes. That is, all images are mapped to labels 0 to 9, e.g., images of digit 0 of MNIST and images of letter $A$ of notMNIST are assigned the label 0. The experiment is illustrated in Figure 6b. In this case, we essentially augment the MNIST dataset with regular and negative images of notMNIST. Similar to the case of training with 20 labels, when tested on negative images of MNIST, the model accuracy reaches $99.8\%$.

Assigning images of the two datasets to the same labels has a practical implication. Most data augmentation techniques include transformed version of images into the train-

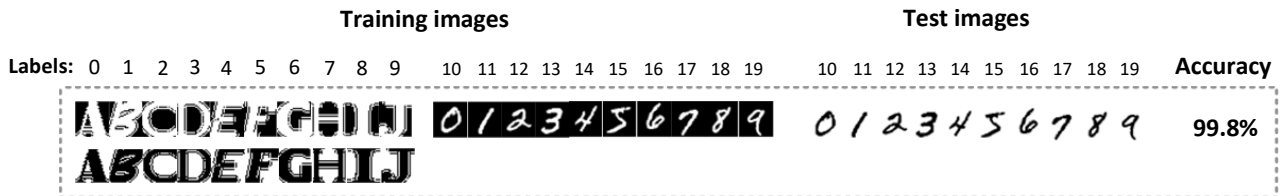

(a) Images of MNIST and notMNIST datasets are mapped to different labels.

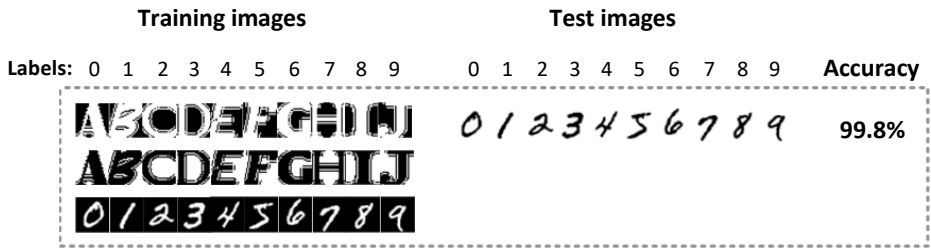

(b) Images of MNIST and notMNIST datasets are mapped to same labels.

Figure 6: An illustration of simultaneously training the model with regular and negative images of notMNIST and regular images of MNIST dataset.

ing data. Our results, however, show that one can augment a target dataset with another dataset that contains similar type of images, though with unrelated labels. Moreover, in the case that the two datasets do not have common classes, each class of the external dataset can be randomly labeled. Augmenting a target dataset with unrelated data is especially helpful, when the dataset size is small and it is difficult to obtain samples from the classes of target dataset.

**Generalization to/from random images.** We examine whether the CNN can learn the shape bias property from a dataset that lacks any structure, e.g., a dataset with random images. We also investigate how a model with shape bias property performs on random images and their negatives. In experiments, we generate a dataset, with the same size as MNIST, consisting of random images with uniform distribution in $[0, 1]$. Images are labeled randomly between 0 to 9.

For answering the first question, we train the sVGG model with random images and their negatives and with the regular images of MNIST. The model is trained with 10 labels, i.e., random images and MNIST images are mapped to the same set of classes. Our experimental results show that the model accuracy on MNIST negative images is about $30\%$, implying that including random images and their negatives in the training data does not improve the model generalization to MNIST negative images. In the second experiment, we train the sVGG model with notMNIST regular and negative images and also with random images. The test accuracy on negatives of random images is $10\%$. In conclusion, although CNN models can fit any dataset, they only learn and generalize the structures.

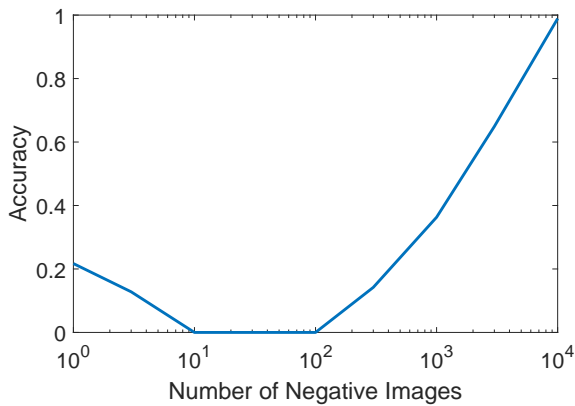

Figure 7: Accuracy of sVGG model on MNIST negative images versus number of notMNIST negative images included in training data. The model is trained with regular and negative images of notMNIST and regular images of MNIST.

### 6.2. Role of Initialization

Now, we examine the effect of initialization on the ability of CNNs to generalize to negative images. For this, we first train the sVGG model with regular and negative images of notMNIST dataset for $100$ epochs, and then finetune with MNIST regular images for another $100$ epochs. We consider two cases for the first phase of training: 1) training with notMNIST regular and negative images with correct labels, 2) training with notMNIST regular and negative images, with labels of negative images being modified to $(i + 1) \bmod 10$, where $i$ is the ground-truth label.

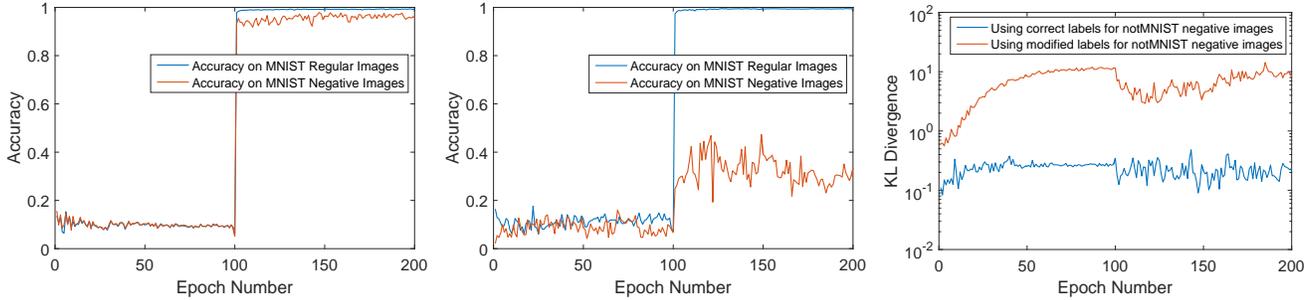

(a) Accuracy versus epoch number in the first case.

(b) Accuracy versus epoch number in the second case.

(c) KL divergence between model outputs on MNIST regular and negative images in both cases.

Figure 8: Role of initialization and fine-tuning on shape bias. The CNN model is trained with notMNIST regular and negative images for 100 epochs and then fine-tuned with MNIST regular images for another 100 epochs. Two cases are considered for the first phase of training: 1) training with notMNIST images with correct labels, and 2) training with notMNIST images, with labels of negative images being modified to $(i+1) \bmod 10$, where $i$ is the ground-truth label. Throughout training, we test the model on MNIST regular and negative images. Only in the first case, the model performs similarly on MNIST regular and negative images, implying that it learns to be invariant to color.

Figures 8a and 8b show the model accuracy on MNIST regular and negative images versus epoch number for the two cases. While accuracy on MNIST images in not meaningful during the first phase of training, notably for the first case, it is similar for both MNIST regular and negative images. This implies that, when correctly trained with regular and negative images of one dataset, the model performs similarly on regular and negative images of another dataset as well. In the first case, after fine-tuning the model with MNIST regular images, the accuracy on negative images increases and remains on par with the accuracy on regular images, i.e., the shape bias property does not fade away when the model is fine-tuned only with regular images. In the second case, the model accuracy on MNIST negative images converges to about $30\%$, which is similar to the case where the model was initialized randomly.

We also use KL-divergence metric to measure the similarity of the model output probability vectors on regular and negative images. Let $P(X)$ be the model output probability vector for an image $X$. The KL-divergence from $P(X)$ to $P(1-X)$ is defined as follows:

$$D_{\text{KL}}(P(X)\|P(1-X)) = \sum_i P(X)_i \log \frac{P(X)_i}{P(1-X)_i}.$$

We define $D = \mathbb{E}_X[D_{\text{KL}}(P(X)\|P(1-X))]$, i.e., the average KL-divergence of model outputs on all pairs of regular and negative images. Figure 8c shows the value of $D$ versus epoch number for the two cases. As can be seen, in the first case (training with correct labels) and during the first phase of training, the KL divergence remains low and on par with the KL divergence in the second phase. This further demonstrates that CNN learns to be invariant to color.

However, in the second case (training with notMNIST negative images with wrong labels), the average KL divergence is significantly higher throughout the training. In conclusion, a different initialization can lead to a qualitatively different model, e.g., in the first case, it is as if the shape bias property is encoded into the model at initialization.

The experiments also show that data augmentation yields better results compared to fine-tuning, as the model seems to fully retain features of the first dataset. Specifically, with fine-tuning and data augmentation, we obtained about $95.4\%$ and $99.8\%$ accuracy on MNIST negative images, respectively. The proposed approach of data augmentation can also mitigate catastrophic forgetting in neural networks [30], a phenomenon in which models "forget" how to perform the first task, when retrained on a second task.

## 7. Conclusion

In this paper, we assessed shape bias property of CNNs, by investigating their ability in generalizing to images with similar shapes but different colors. In experiments, we used original and negative images of training data as such images. Through experiments, we evaluated the role of various components of training algorithms in generalizability. We showed that CNNs do not display shape bias by design. They, however, can learn to classify objects based on their shapes, when the model is properly initialized or data is properly augmented, and if batch normalization is used.

## Acknowledgments

This work was supported by ONR grants N00014-14-1-0029 and N00014-16-1-2710, ARO grant W911NF-16-1-0485 and NSF grant CNS-1446866.